\documentclass{article}

\usepackage[english]{babel}

\usepackage[letterpaper,top=2cm,bottom=2cm,left=3cm,right=3cm,marginparwidth=1.75cm]{geometry}

\usepackage{graphicx}
\usepackage[colorlinks=true, allcolors=blue]{hyperref}

\usepackage{lipsum}
\usepackage{amsmath}
\usepackage{amssymb}
\usepackage{mathtools}
\usepackage{amsthm}
\usepackage{tikz} % nice language for creating drawings and diagrams
\usepackage{amsfonts,bm}
\usepackage{booktabs}
\usepackage{enumitem}
\usepackage{graphicx}
\usepackage{subfigure}
\usepackage{cite}
\usepackage[capitalize,noabbrev]{cleveref}

\title{GPgym: A Remote Service Platform \\ with Gaussian Process Regression for Online Learning}
\author{Xiaobing Dai, Zewen Yang}
\date{}
\begin{document}
\maketitle

Machine learning is now widely applied across various domains, including industry, engineering, and research. While numerous mature machine learning models have been open-sourced on platforms like GitHub, their deployment often requires writing scripts in specific programming languages, such as Python, C++, or MATLAB. This dependency on particular languages creates a barrier for professionals outside the field of machine learning, making it challenging to integrate these algorithms into their workflows.

To address this limitation, we propose GPgym in \cite{anonymous2024asynchronous,yang2024asynchronousdistributedgaussianprocess}, a remote service node based on Gaussian process regression. GPgym enables experts from diverse fields to seamlessly and flexibly incorporate machine learning techniques into their existing specialized software, without needing to write or manage complex script code.

GPgym leverages Gaussian process regression, a non-parametric learning approach capable of modeling any continuous function with high accuracy. This flexibility makes it particularly suited for a wide range of applications.

In the following sections, we will provide detailed instructions on the installation, functionality, and setup of GPgym, demonstrating how it empowers users to solve professional problems with machine learning in a user-friendly manner.

\section{Installation}

GPgym is a MatLab-based software, which can be downloaded from: 

\url{https://github.com/Xiaobing-Dai/GPgym}
\\

 \noindent  Note that GPgym requires \textit{MATLAB Runtime} environment corresponding to MatLab 2021b.
\textit{MATLAB Runtime} can be downloaded from the following link:

\url{https://ww2.mathworks.cn/products/compiler/matlab-runtime.html}

\section{Functions}

\subsection{Configuration of Gaussian Process}

To provide computationally efficient learning and inference without loss of data information, a variant of GP is applied, namely the locally growing random tree of GPs (LoG-GP, \cite{pmlr-v139-lederer21a}). 
LoG-GP separates the data set into several local GP models in a tree structure. 
In GPgym, we provide a user interface to decide:
\begin{itemize}
	\item the maximal number of local GP models
	\item and the maximal number of data samples in each local GP model.
\end{itemize}

For each local GP model, we use automatic relevance determination squared exponential (ARD-SE) kernel, which has the expression as
\begin{align}
	k(\bm{x}, \bm{x}') = \sigma_f^2 \exp \left( - \frac{1}{2} \sum_{d=1}^D \frac{(x_d - x'_d)^2}{l_d^2} \right)
\end{align}
with $\bm{x} = [x_1, \cdots, x_D]^T \in \mathbb{R}^D$ and $\bm{x}' = [x'_1, \cdots, x'_D]^T \in \mathbb{R}^D$, where $D \in \mathbb{N}_+$ is the dimension of the input $\bm{x}$.
The hyperparameters, i.e., 
\begin{itemize}
	\item $\sigma_f \in \mathbb{R}_+$ -- variance,
	\item $l_d \in \mathbb{R}_+, \forall d = 1, \cdots, D$ -- length scale
\end{itemize}
are selected by users.

Moreover, the information of data quality is essential for GP prediction, which is characterized by
\begin{itemize}
	\item $\sigma_n \in \mathbb{R}_+$ -- variance of measurement noise.
\end{itemize}

All the above-mentioned parameters can be configured by the user interface in GPgym.

\subsection{Remote Services}

GPgym uses UDP ports to provide remote services. Specifically, GPgym utilizes 2 UDP ports for receiving and sending data, whose configuration data including
\begin{itemize}
	\item IP address for UDP reading port,
	\item port number for UDP reading port,
	\item IP address for UDP sending port,
	\item port number for UDP sending port
\end{itemize}
are set in the user interface.

When data is received at the UDP reading port, GPgym reacts differently based on the received data form. Specifically:
\begin{itemize}
	\item If the received data is a scalar, the initialization of LoG-GP will be activated.
	\item If the received data is a vector with at least $3$ dimensions, the online learning will be activated. 
\end{itemize}

In detail for the online learning case, the received data $\bm{p} \in \mathbb{R}^{m}$ with $m \in \mathbb{N}_{\ge 3}$ is separated as 
\begin{align}
	\bm{p} = [\bm{x}^T, y, t]^T
\end{align}
where $\bm{x}$ is the inquiry point, $y$ is the measured value of unknown function at $\bm{x}$ and $t$ denotes the time stamp.
The data pair $\{ \bm{x}, y \}$ will be added into the data set of LoG-GP, and then provides the inference of unknown function at $\bm{x}$, denoted as $\mu(\bm{x})$.
The generated prediction $\mu(\bm{x})$ will be sent back combined with the time stamp $t$ via UDP sending port.

More detailed instructions on how to configure LoG-GP and start the remote service are shown in the next section.

\section{GPgym Instruction}\label{sec_gymInstruction}
In this section, we provide a comprehensive set of instructions for utilizing OLGPgym, encompassing three key steps. These steps include setting up User Datagram Protocol (UDP), configuring GP models and datasets, and executing simulations. The following subsections offer detailed operational guidance with an intuitive user interface.

%%%%%%%%%%%%%%%%%%%%%%%%%%%%%%%%%%%%%%%%%%%%%%%%%%%%%%%%%%%%%%%%%%%%%%%%%
\subsection{Set Up UDP}
\begin{figure}[ht]
    \centering
    \includegraphics[scale=0.2]{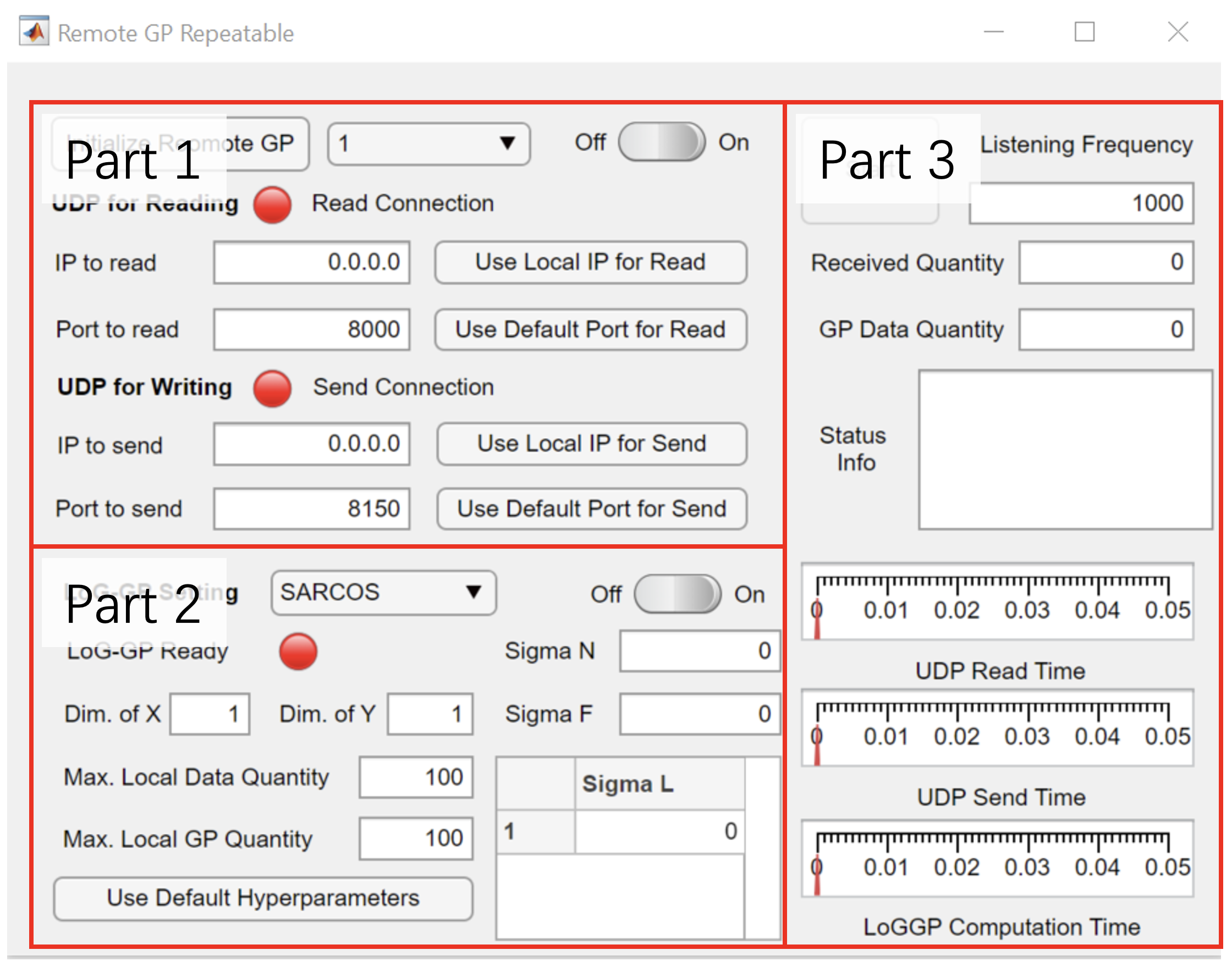}
    \caption{GUI.} 
    \label{fig_UGI_intro}
\end{figure}

After executing the \texttt{RemoteGP.exe}, the MATLAB App is open. The graphical user interface (GUI) is composed of 3 parts shown in \cref{fig_UGI_intro}.

% \begin{wrapfigure}{r}{0.5\textwidth}
%   \begin{center}
%     \includegraphics[scale=0.18]{fig/GUI_IP.png}
%   \end{center}
%   \caption{Birds}
%   \label{fig_GUI_IP}
% \end{wrapfigure}

The initial phase involves configuring communication settings utilizing the UDP. To establish UDP communication, it is imperative to specify the IP address and ports designated for information transmission and reception (refer to the red box in Figure 2). 
The components ``IP to read'' and ``Port to read'' are designed to receive character and numerical values, respectively, representing the address for information, such as system states and measurements, dispatched from the server. The button ``Use Local IP for Read'' automatically sets ``IP to read'' to the current IPV4 address assigned to the running computer. The ``Use Default Port for Read'' sets "Port to read" to 8000, a predefined port for the server to transmit data.
\begin{figure}[ht]
\centering
\begin{minipage}[c]{0.43\linewidth}
\label{fig_GUI_reading}\includegraphics[width=\linewidth]{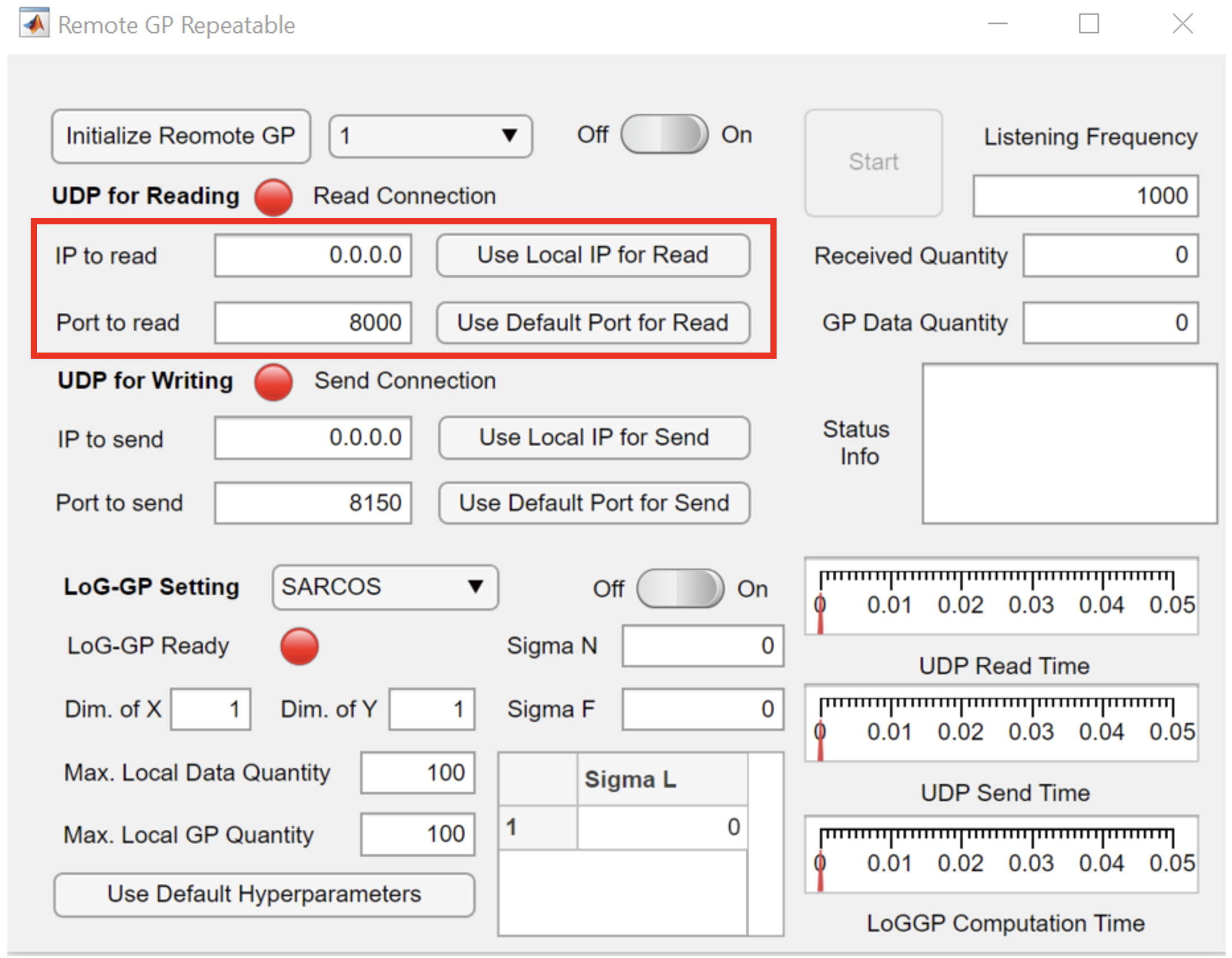}
\caption{UDP for reading.}
\end{minipage}
\hspace{10pt}
\begin{minipage}[c]{0.43\linewidth}
\label{fig_GUI_writing}\includegraphics[width=\linewidth]{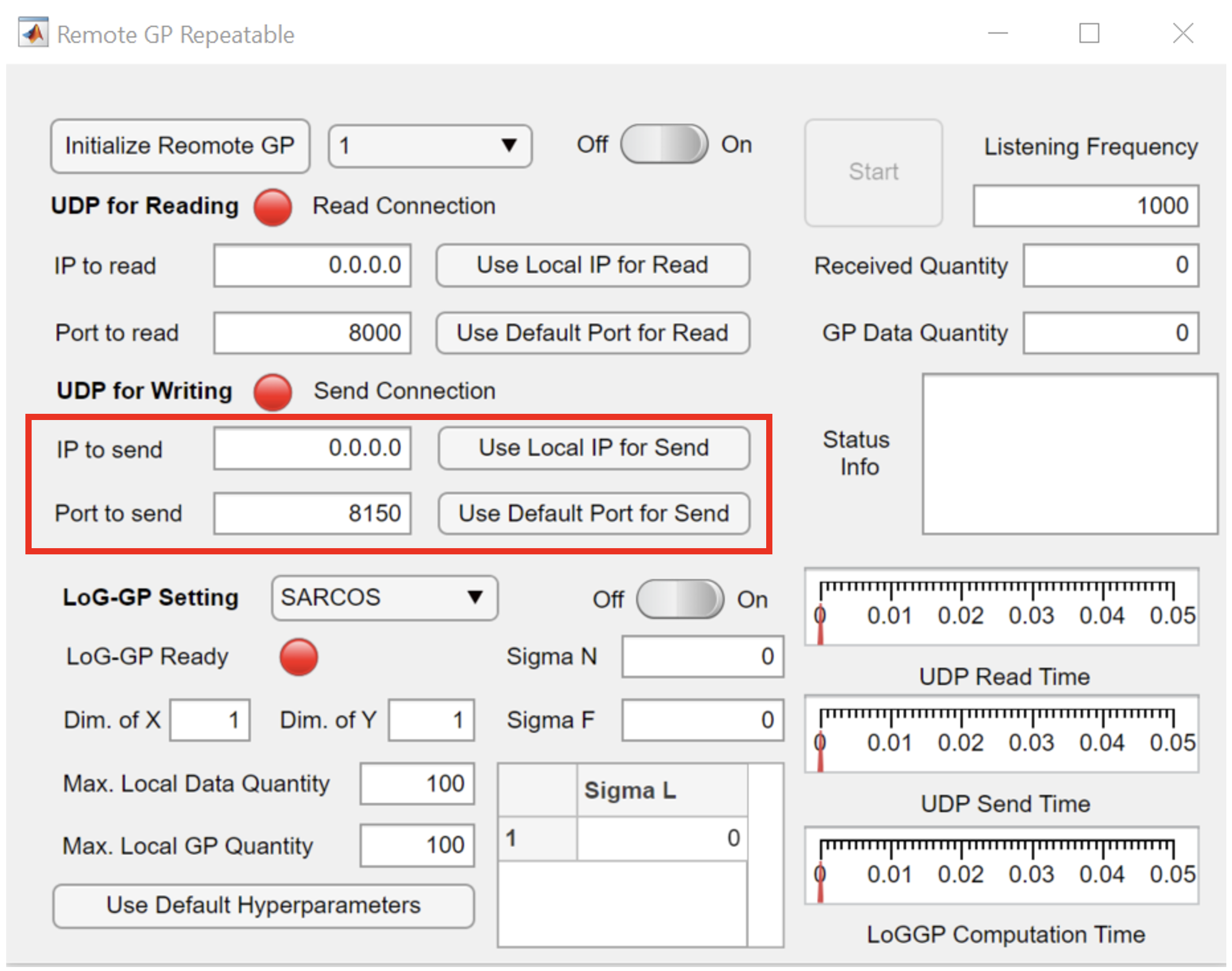}
\caption{UDP for writing.}
\end{minipage}%
\end{figure}

Likewise, configuring UDP for sending results to the server is essential. This setup involves specifying the IP and Port in the ``IP to send'' and ``Port to send'' fields, typically corresponding to the IP address and port on the server. The "Use Local IP for Send" button sets ``IP to read'' to the current IPV4 address assigned to the running computer in Figure 3. Additionally, the ``Use Default Port for Send'' sets ``Port to read'' to 8050, serving as the default port.

%%%%%%%%%%%%%%%%%%%%%%%%%%%%%%%%%%%%%%%%%%%%%%%%%%%%%%%%%%%%%%%%%%%%%%%%%
\subsection{Configuring GP Models and Datasets}
\begin{figure}[ht]
    \centering
    \includegraphics[scale=0.2]{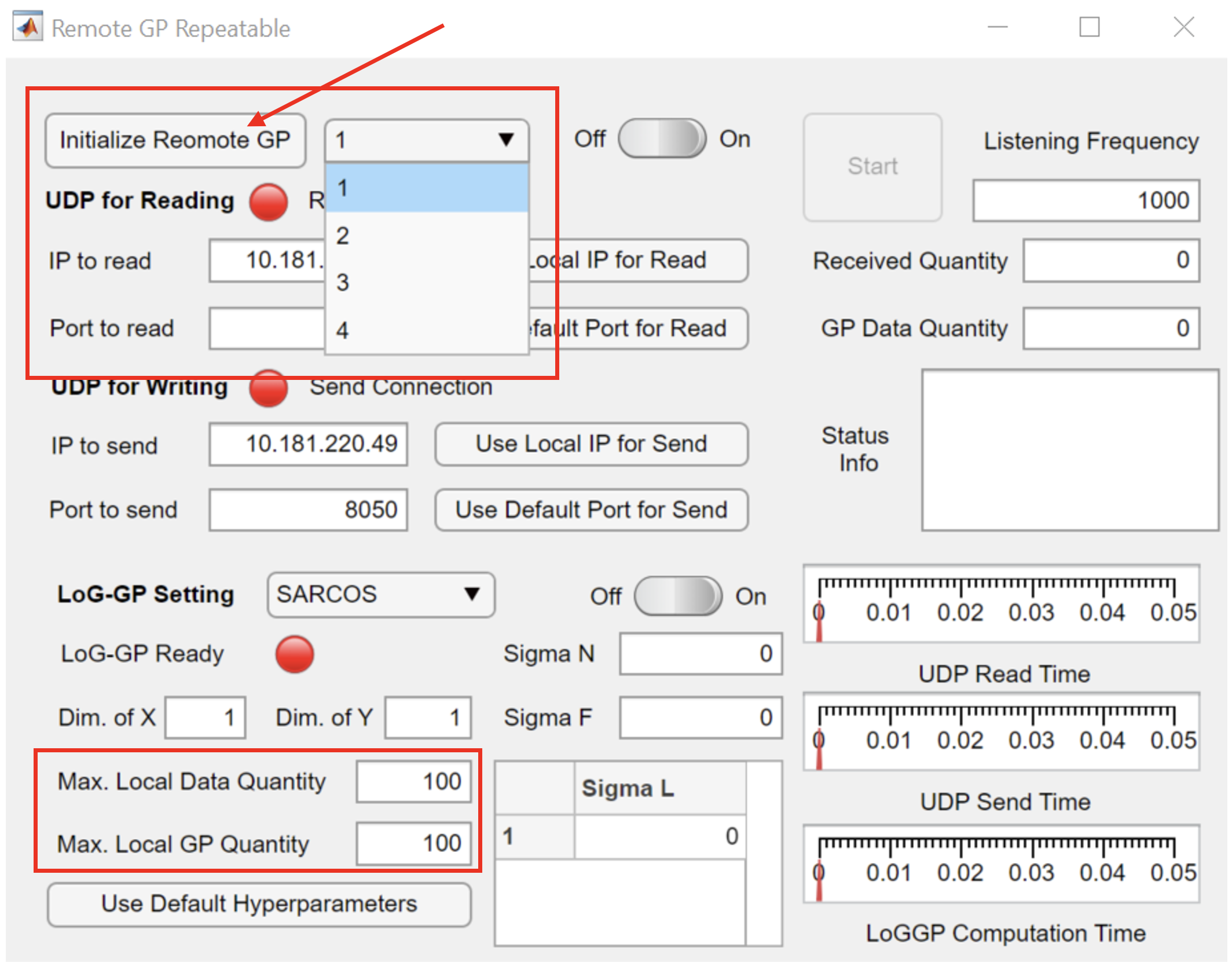}
    \caption{Initialize GP models.}
    \label{fig_GUI_initialGP}
\end{figure}

Before turning on a remote GP model, several predefined settings are available for users. By selecting a specific model (the default configuration includes 4 models) and pressing the ``Initialize Remote GP'' button, the associated IP and ports for reading and sending will be automatically populated in the ``IP to read'', ``Port to read'', ``IP to send'', and ``Port to send'' fields (see \cref{fig_GUI_initialGP}). Furthermore, the default settings provide values for the maximal number of tree leaves and ``Max. Local Data Quantity'' meaning the maximal number of data in each leaf node, which are the parameters for the applied LoG-GP model, with the option for customization within the code script. 

\begin{figure}[ht]
    \centering
    \includegraphics[scale=0.2]{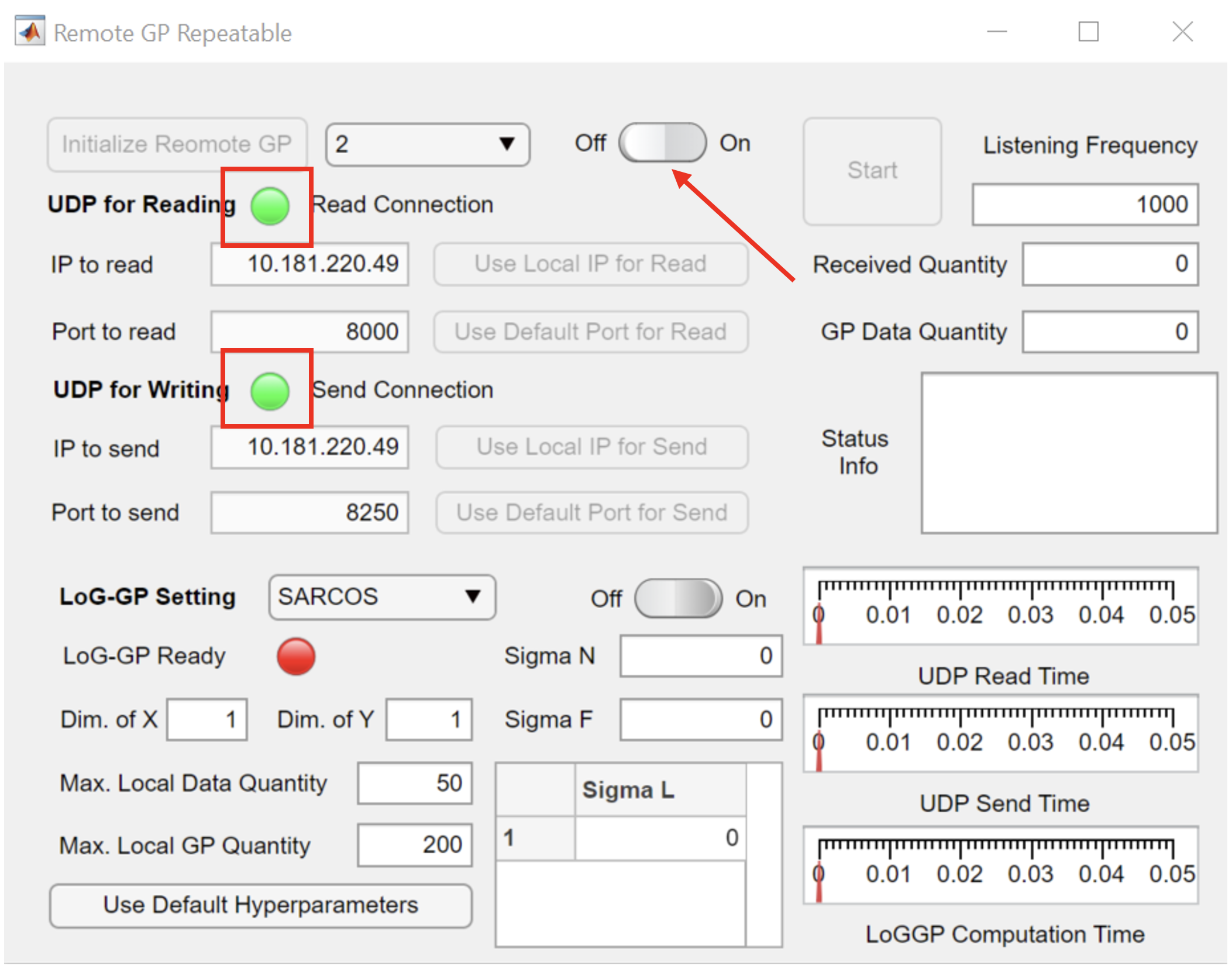}
    \caption{Activate UDP communication.}
    \label{fig_GUI_turnOn}
\end{figure}

Once the IP and ports for UDP reading and sending are configured, initiate UDP communication by turning on the switch. Successful opening of UDP ports will be indicated by the turning of the index light to green (see \cref{fig_GUI_turnOn}). In the event of an error, the lights will remain red, signaling an issue. In such cases, deactivate UDP, review the UDP send and read settings, ensuring that the server and remote GP are on the same network and that the ports are not occupied. After verification, re-activate the switch. During the initial run of the GUI, certain warnings may arise. In such instances, authorizing the application to establish network connections is needed.

Furthermore, it is imperative to specify the dimensions of both the input and output for the GP. Multi-dimensional GPs are supported, allowing for training output data and prediction to possess multiple dimensions.
In the context of GPs, the selection of hyperparameters is crucial. In this implementation, the automatic relevance determination squared exponential (ARD-SE) kernel is employed. This kernel necessitates defining parameters such as "Sigma F" for overall variance, "Sigma L" for length scales for each input dimension, and "Sigma N" for the variance associated with measurement noise.
Upon specifying the training input data dimension, the corresponding fill area for "Sigma L" is automatically generated.
Additionally, predefined settings for GP are available for different datasets, including dimensions for input and out along with hyperparameters. These datasets include SARCOS, KIN40K, POL, PUMADYN32NM, Control, and Toy.
After selecting the desired predefined setting, pressing ``Use Default Hyperparameters'' loads the values from the associated \texttt{.mat} file. Consequently, all dimensions and hyperparameters necessary for the GP are applied accordingly.
\begin{figure}[ht]
    \centering
    \includegraphics[scale=0.2]{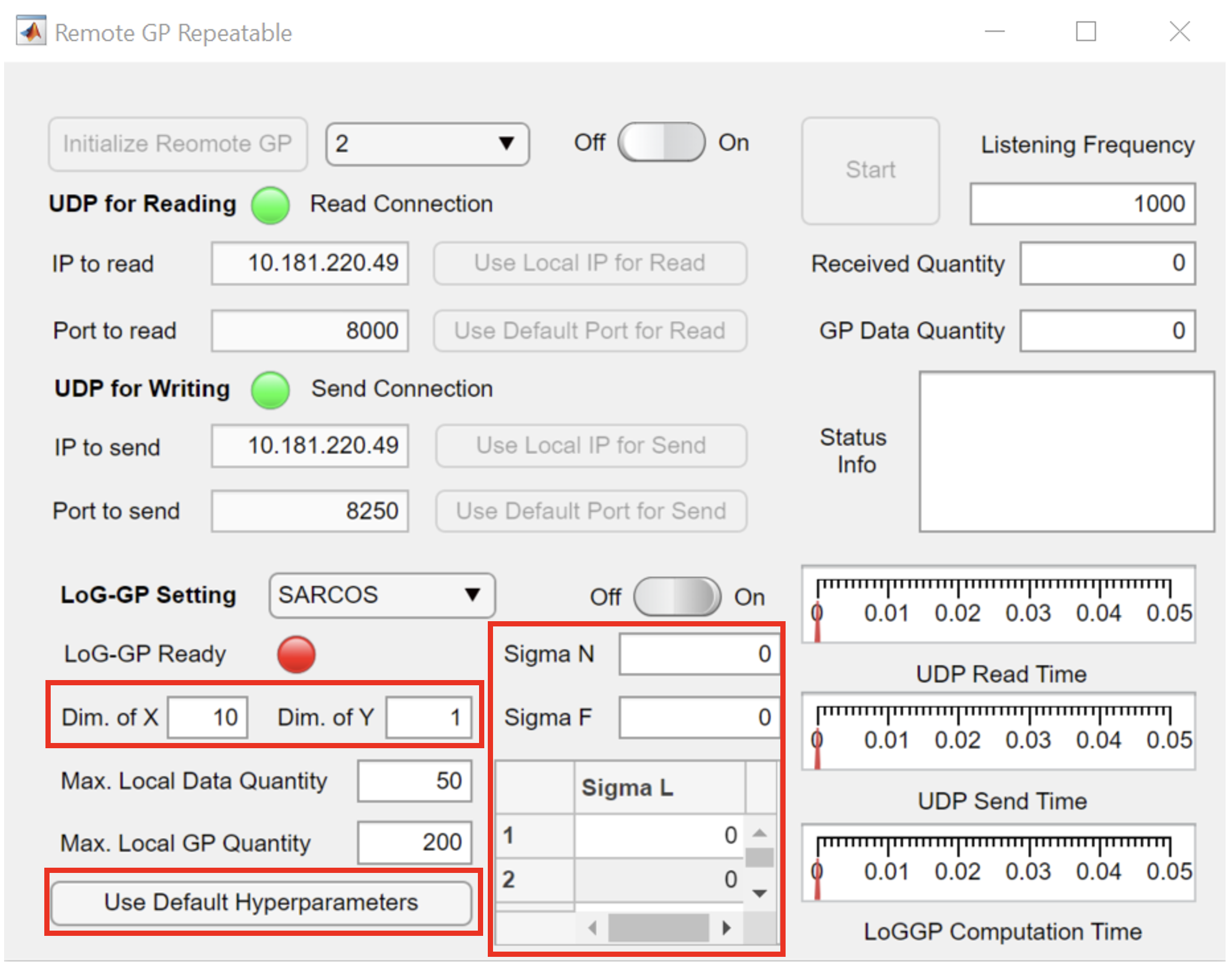}
    \caption{Set up GP models' parameters.}
    \label{fig_GUI_params}
\end{figure}

\begin{figure}[ht]
    \centering
    \includegraphics[scale=0.2]{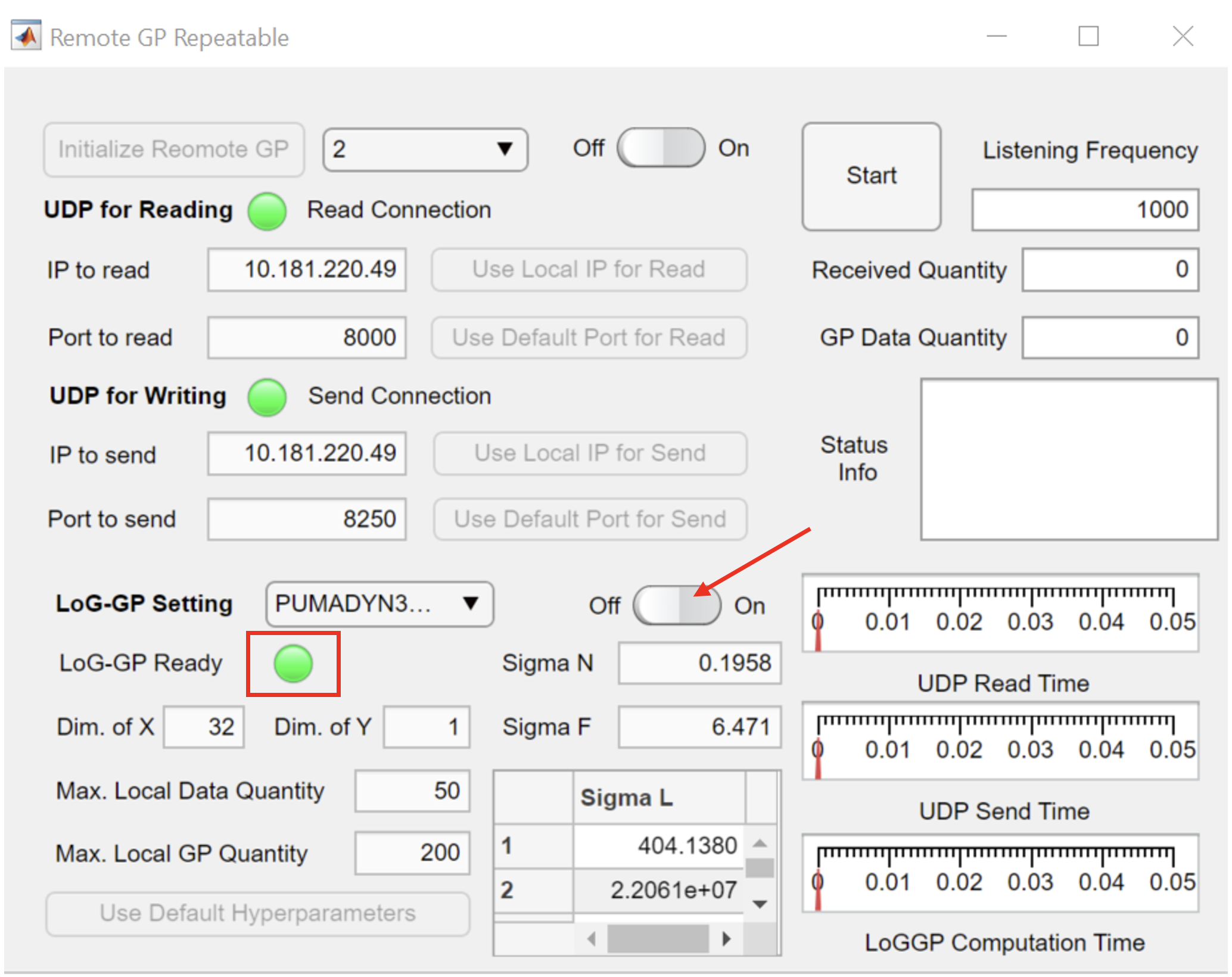}
    \caption{Activate GP models.}
    \label{fig_GUI_GPTurnON}
\end{figure}

After setting up the GP model, activating the switch results in the creation of a new empty GP object. This process concurrently implies the deletion of any existing GP object, if present. The change in the light to green signifies that the GP model is now in a standby state in \cref{fig_GUI_GPTurnON}.
It is noteworthy that a successful GP model creation leads to the locking of operations for the ``Use Default Hyperparameters'' button and the input area for dimensions and hyperparameters.

\subsection{Executing Simulations}
The third part is the operation of running the remote GP and providing feedback on its running status. Once both UDP and GP configurations are successfully set up, the application is prepared for execution. 
The only setting area in this part is the listening frequency of UDP read. 
After configuring the frequency, pressing the ``Start'' button initiates the operation of the remote GP.
\begin{figure}[ht]
\centering
\begin{minipage}[c]{0.43\linewidth}
\includegraphics[width=\linewidth]{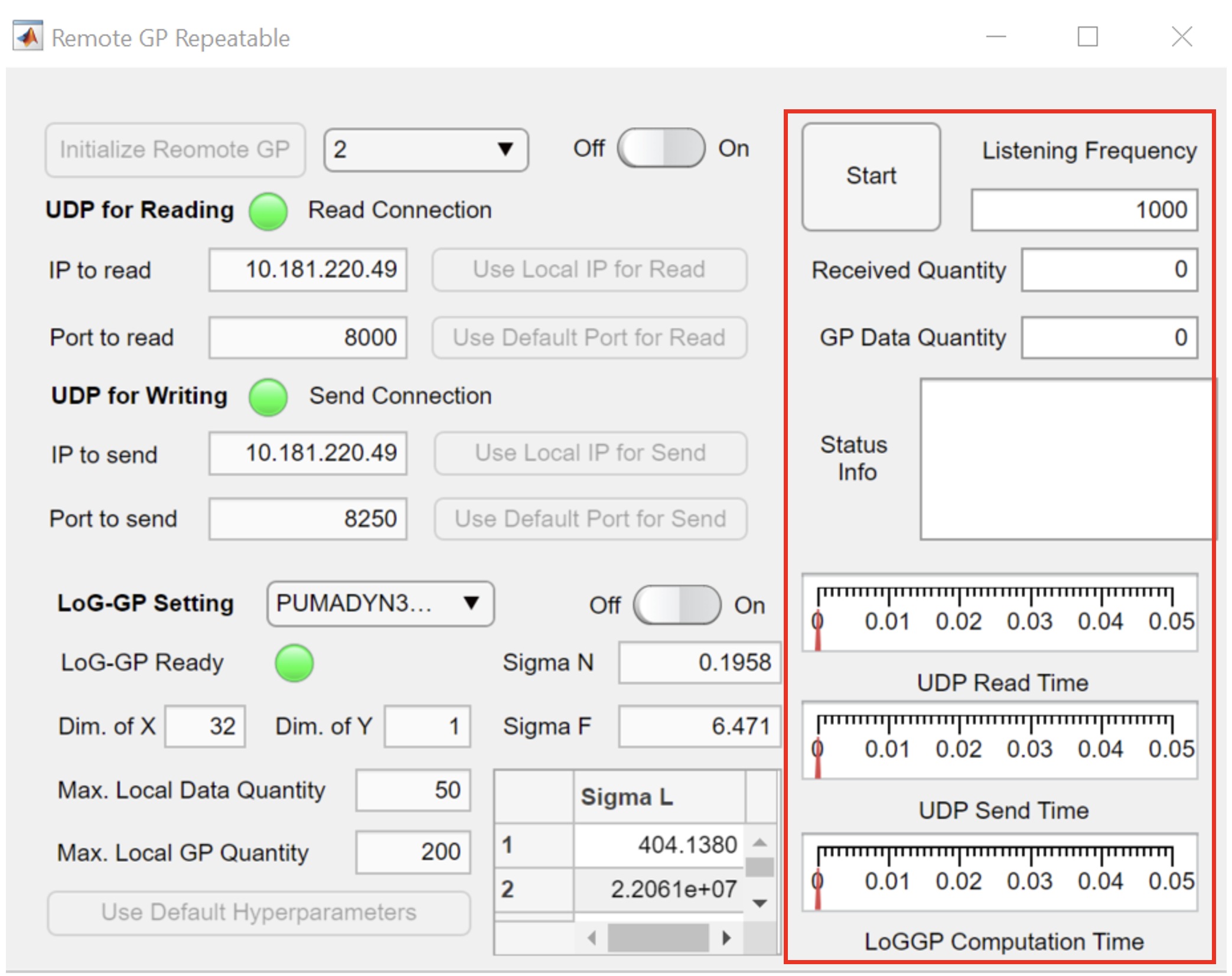}
\caption{The third part of GUI.}
\end{minipage}
\hspace{10pt}
\begin{minipage}[c]{0.43\linewidth}
\includegraphics[width=\linewidth]{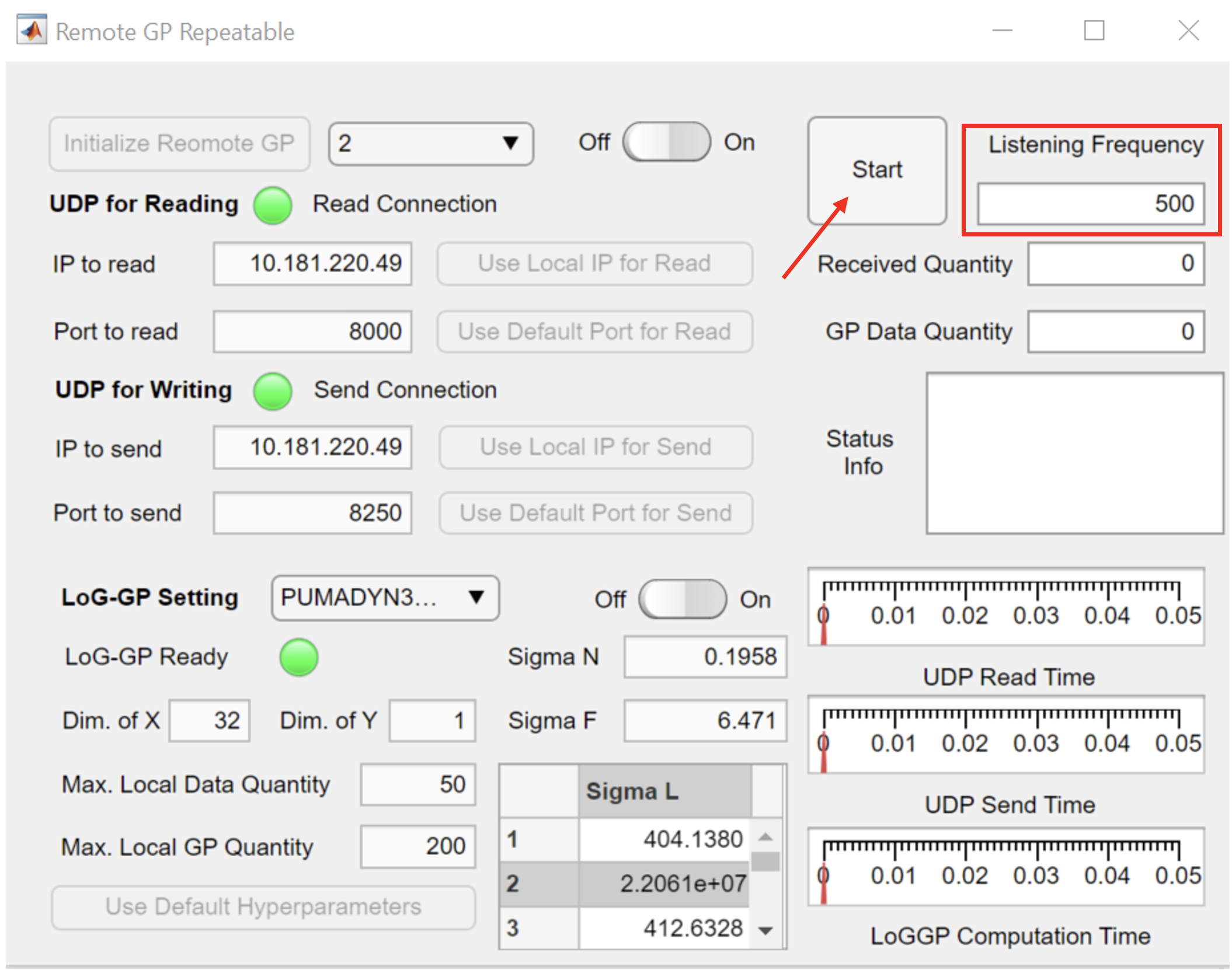}
\caption{Start GP model.}
\end{minipage}%
\end{figure}

\begin{figure}[ht]
    \centering
    \includegraphics[scale=0.2]{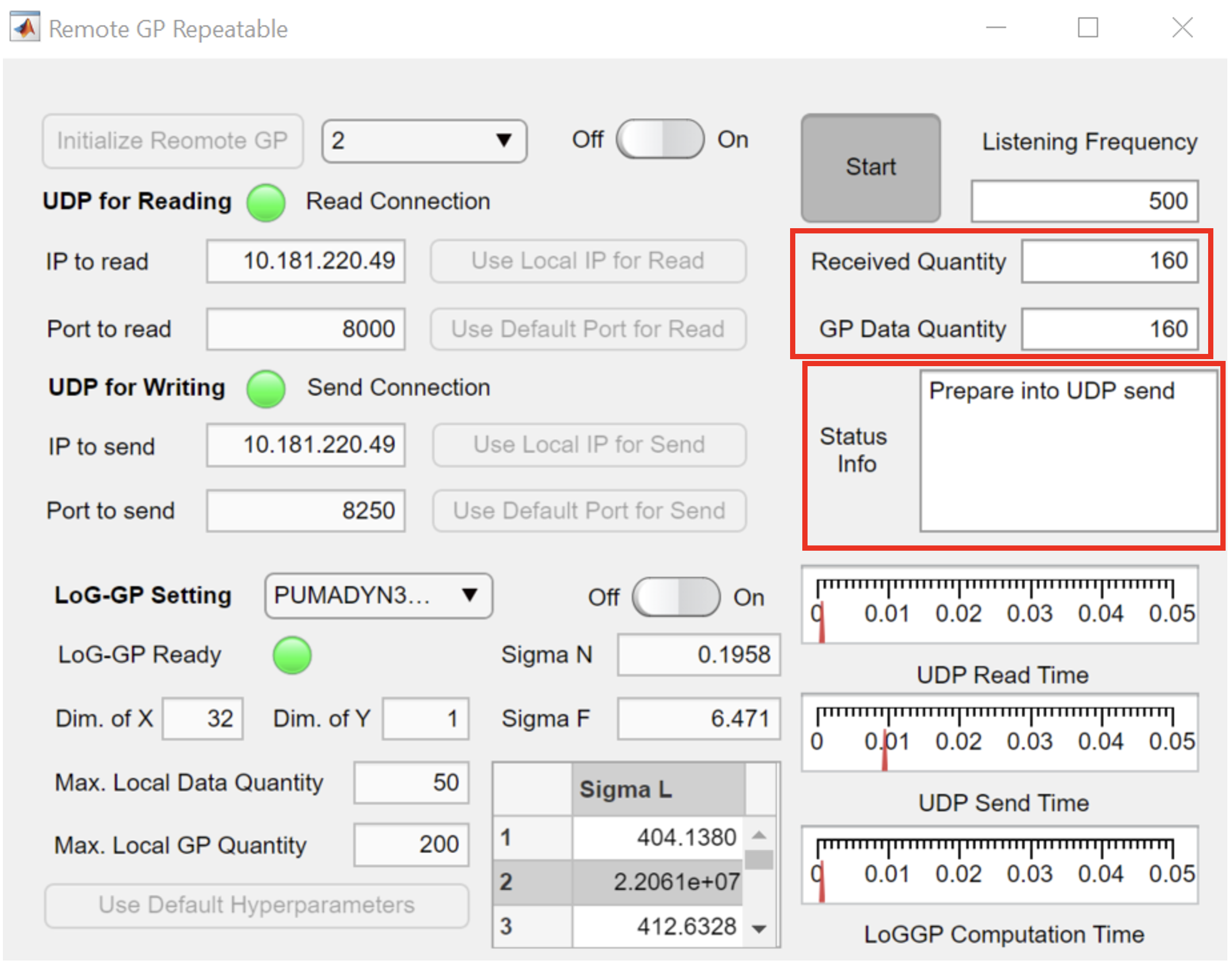}
    \caption{Information of GP models after start.}
    \label{fig_GUI_info}
\end{figure}

The areas labeled "Received Quantity" and "GP Data Quantity" provide information on the number of data received from the server and the quantity of stored training data sets in the GP model, respectively (see \cref{fig_GUI_info}). It's worth noting that these two values may differ due to the limited data storage defined by parameters such as ``Max. Local Data Quantity'' and ``Max. Local GP Quantity''.
Three gauges are deployed to visually represent the time taken for different tasks. Specifically, the focus is on communication time and computation time. This encompasses the time taken for UDP to read data from the server, the time taken for UDP to send results to the server, and the time consumed by the GP model for online learning and prediction.
The ongoing operation will persist until a specific command is received or until manually terminated by pressing the "Start" button again.
The special command is denoted by a single value "-1" received from the server. Upon receiving this value, the GP model will be initialized with an empty data set. This feature can be leveraged for conducting Monte Carlo tests. It is important to note that the deactivation of either the UDP or GP model switch will result in the termination of the remote GP operation. 

\begin{figure}[ht]
    \centering
    \includegraphics[width=\textwidth]{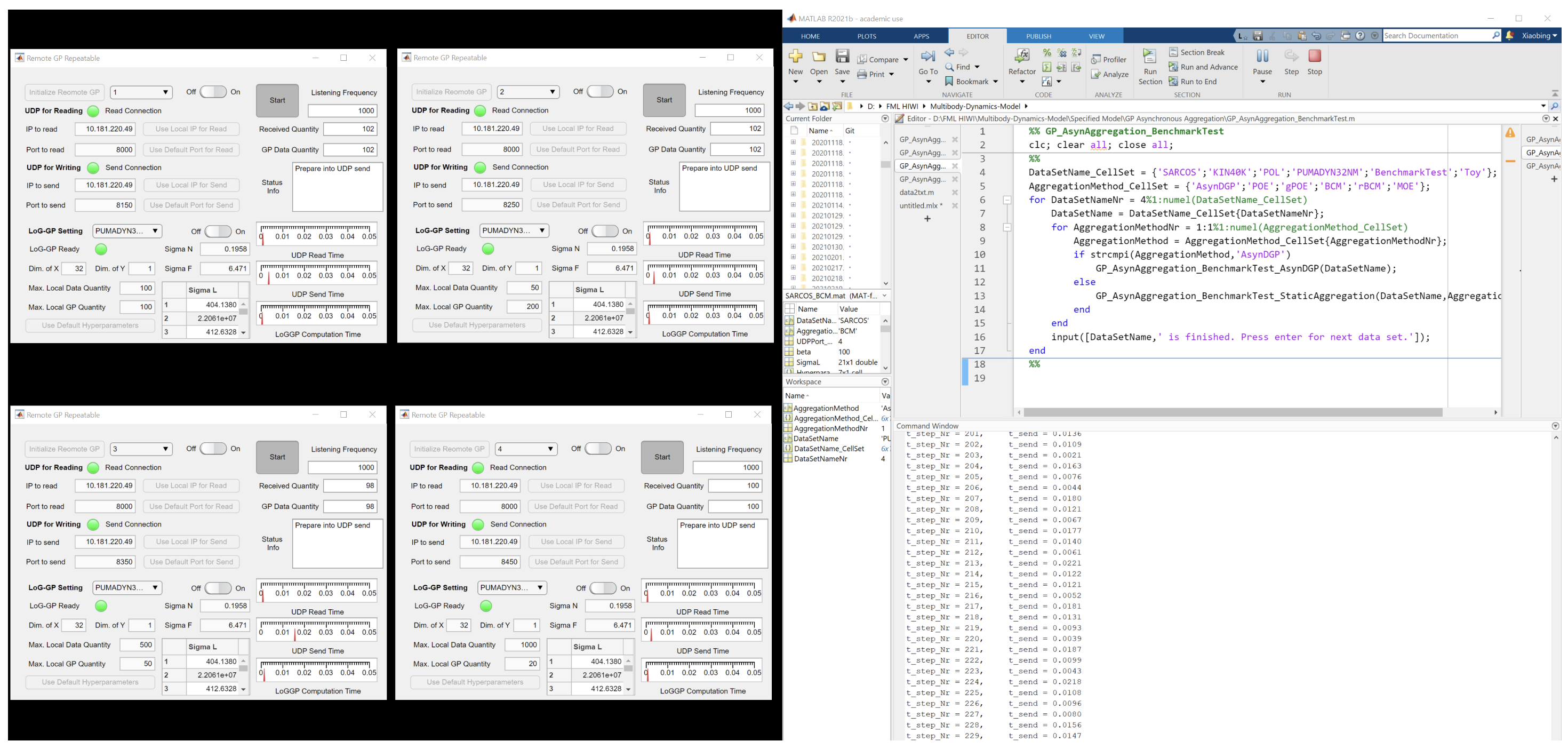}
    \caption{Execute simulations.}
    \label{fig_experiment}
\end{figure}

After all GP models are activated, we are able to execute the simulations of either regression tasks or control tasks with scripts, which is shown in \cref{fig_experiment}.

\bibliography{main}

\begin{thebibliography}{1}

\bibitem{anonymous2024asynchronous}
Z.~Yang, X.~Dai, and S.~Hirche, ``{Asynchronous Distributed Gaussian Process
  Regression},'' in {\em The 39th Annual AAAI Conference on Artificial
  Intelligence}, 2024.

\bibitem{yang2024asynchronousdistributedgaussianprocess}
Z.~Yang, X.~Dai, and S.~Hirche, ``{Asynchronous Distributed Gaussian Process
  Regression for Online Learning and Dynamical Systems: Complementary
  Document},'' 2024.

\bibitem{pmlr-v139-lederer21a}
A.~Lederer, A.~J.~O. Conejo, K.~A. Maier, W.~Xiao, J.~Umlauft, and S.~Hirche,
  ``{Gaussian Process-Based Real-Time Learning for Safety Critical
  Applications},'' in {\em Proceedings of the 38th International Conference on
  Machine Learning} (M.~Meila and T.~Zhang, eds.), vol.~139 of {\em Proceedings
  of Machine Learning Research}, pp.~6055--6064, PMLR, 18--24 Jul 2021.

\end{thebibliography}
\bibliographystyle{ieeetr}

\end{document}